\pdfoutput=1

\documentclass[11pt]{article}

\usepackage{ACL2023}

\usepackage{times}
\usepackage{latexsym}

\usepackage[T1]{fontenc}

\usepackage[utf8]{inputenc}

\usepackage{microtype}

\usepackage[export]{adjustbox}
\usepackage{inconsolata}
\usepackage{times}
\usepackage{latexsym}
\usepackage{enumitem}
\usepackage{multirow}
\usepackage{array}
\usepackage{amsfonts}
\usepackage{subcaption}
\usepackage{makecell}
\usepackage{soul}
\usepackage{xcolor,colortbl}
\usepackage{xspace}

\usepackage{arydshln}
\usepackage{textcomp}
\usepackage{stfloats}
\usepackage{url}
\usepackage{verbatim}
\usepackage{graphicx}
\usepackage{algorithm}
\usepackage{algorithmic}
\usepackage{amssymb}
\usepackage{helvet}  
\usepackage{courier}  
\usepackage{newfloat}
\usepackage{listings}
\usepackage{multirow}
\usepackage{amsmath}
\usepackage{booktabs}
\usepackage{pifont}
\usepackage{color}

\newcommand{{\model}}{HugNLP}

\definecolor{codegreen}{rgb}{0,0.6,0}
\definecolor{codegray}{rgb}{0.5,0.5,0.5}
\definecolor{codepurple}{rgb}{0.58,0,0.82}
\definecolor{backcolour}{rgb}{100, 100, 100}

\lstdefinestyle{mystyle}{
    backgroundcolor=\color{backcolour},   
    commentstyle=\color{codegreen},
    stringstyle=\color{codepurple},
    basicstyle=\ttfamily\scriptsize,
    breakatwhitespace=true,         
    breaklines=true,                 
    captionpos=b,                    
    keepspaces=true,                 
    numbers=none,                    
    numbersep=5pt,                  
    showspaces=false,                
    showstringspaces=false,
    showtabs=false,                  
    tabsize=2,
    columns=flexible,
    escapeinside={(*}{*)},
}

\title{HugNLP: A Unified and Comprehensive Library for \\ Natural Language Processing}

\author{Jianing Wang$^1$, Nuo Chen$^1$, Qiushi Sun$^{1,4}$, Wenkang Huang$^2$, Chengyu Wang$^{3}$, Ming Gao$^{1}$\thanks{ \quad Corresponding Author.}\\
  $^1$ School of Data Science and Engineering, East China Normal University, Shanghai, China \\
  $^2$ Ant Group, Shanghai, China $^3$ Alibaba Group, Hangzhou, China\\
  $^4$ National University of Singapore, Singapore\\
  \texttt{lygwjn@gmail.com,} \texttt{\{nuochen,qiushisun\}@stu.ecnu.edu.cn} \\
  \texttt{\{wenkang.hwk,chengyu.wcy\}@alibaba-inc.com, mgao@dase.ecnu.edu.cn}
}

\begin{document}
\maketitle
\begin{abstract}
In this paper, we introduce~{\model}, a unified and comprehensive library for natural language processing (NLP) with the prevalent backend of HuggingFace Transformers, which is designed for NLP researchers to easily utilize off-the-shelf algorithms and develop novel methods with user-defined models and tasks in real-world scenarios.
{\model} consists of a hierarchical structure including models, processors and applications that unifies the learning process of pre-trained language models (PLMs) on different NLP tasks.
Additionally, we present some featured NLP applications to show the effectiveness of {\model}, such as knowledge-enhanced PLMs, universal information extraction, low-resource mining, and code understanding and generation, etc.
The source code will be released on GitHub (\url{https://github.com/wjn1996/HugNLP}).

\end{abstract}

\section{Introduction}

Recently, pre-trained language models (PLMs) have become the imperative infrastructure in a series of downstream natural language processing (NLP) tasks~\cite{Devlin2019BERT, Liu2019RoBERTa, Yang2019XLNet}, which bring substantial improvements by a two-stage training strategy, including \emph{pre-training} and \emph{fine-tuning}.
Benefiting from this strategy, a branch of PLM methods arises to improve the models' effectiveness, promoting NLP's development in both academia and industry \cite{Liu2023Pre, Hu2022A}.

Yet, many existing approaches follow different patterns and code architectures, it is not easy to obtain high-performing models and develop them easily for researchers. 
To fill this gap, this paper presents {\model}, a unified and comprehensive open-source library to allow researchers to develop and evaluate NLP models more efficiently and effectively.
To reach this goal, we utilize HuggingFace Transformers~\footnote{\url{https://huggingface.co/}.} as the prevalent backend, which provides abundant backbones of different scale-sizes of PLMs.
For training, we integrate a well-designed tracking toolkit \emph{MLFlow}~\footnote{\url{https://www.mlflow.org/}.} into the backend, which is convenient to observe experimental progress and records.  
{\model} consists of some well-designed components, such as \emph{Models}, \emph{Processors}, and \emph{Applications}.
Concretely,
1) for \emph{Models}, we provide some popular PLMs, including BERT~\cite{Devlin2019BERT}, RoBERTa~\cite{Liu2019RoBERTa}, DeBERTa~\cite{He2021Deberta}, GPT-2~\cite{radford2019language} and T5~\cite{Raffel2020Exploring}, etc.
Based on these PLMs, we develop task-specific modules for pre-training (e.g., masked language modeling (MLM), casual language modeling (CLM)) and fine-tuning (e.g., sequence classifying and matching, span extraction, text generation).
We also provide some prompt-based fine-tuning techniques which enable parameter-efficient tuning for PLMs, including PET~\cite{Schick2021Exploiting}, P-tuning~\cite{Liu2021GPT}, Prefix-tuning~\cite{Li2021Prefix}, Adapter-tuning~\cite{Houlsby2019Parameter}.
2) In \emph{Processors}, we develop relevant data processing tools~\footnote{The \emph{Processor} is related to the task format. For example, we tailor some benchmark datasets, such as Chinese CLUE~\cite{Xu2020CLUE}, GLUE~\cite{Wang2019GLUE}, etc.
}
for some commonly used benchmark datasets and business-specific corpora.
3) In \emph{Applications}, we present core capacities to support the upper applications.
Specifically, our proposed KP-PLM~\cite{Wang2022Knowledge} enables plug-and-play knowledge injection in model pre-training and fine-tuning via converting structure knowledge into unified language prompts.
We also develop HugIE, a universal information extraction toolkit through instruction-tuning with extractive modeling (e.g., global pointer) \cite{Su2022Su}.
{\model} also integrates some novel algorithms and applications, such as uncertainty-aware self-training \cite{Mukherjee2020Uncertainty, Wang2023Uncertainty}, code understanding and generation~\cite{feng2020codebert, Wang2021CodeT5}.

Overall, {\model} has the following features.
\begin{itemize}
    \item {\model} offers a range of pre-built components and modules (i.e., \emph{Models}, \emph{Processors}, \emph{Applications}) that can be used to speed up the development process and simplify the implementation of complex NLP models and tasks. 
    
    \item {\model} can also be easily integrated into existing workflows and customized to meet the specific needs of individual researchers or projects, ensuring the framework's scalability and flexibility.

    \item {\model} is equipped with some novel core capacities, such as knowledge-enhanced pre-training, prompt-based fine-tuning, instruction and in-context learning, uncertainty-aware self-training, and parameter-efficient learning. We thus develop some featured products or solutions
    on real-world application scenarios, e.g., KP-PLM, and HugIE.

    \item HugNLP is based on PyTorch and HuggingFace, which are two widely used tools and platforms in the NLP community, allowing researchers to leverage their strengths and applying it to different academics and industry scenarios~\cite{Qiu2021EasyTransfer,Wang2022EasyNLP}.

\end{itemize}

\section{Background}

\subsection{Pre-trained Language Models}

The goal of the PLM is to learn semantic representations over unsupervised corpora via well-designed self-supervised learning tasks in the pre-training stage. 
Notable PLMs can be divided into three main types, including encoder-only~\cite{Devlin2019BERT, Liu2019RoBERTa, He2021Deberta, Yang2019XLNet, Lan2020ALBERT}, decoder-only~\cite{radford2018improving, Brown2020Language, Zhang2022OPT} and encoder-decoder~\cite{Lewis2020BART, Raffel2020Exploring}.
However, these PLM may lack of background knowledge when applied to some task-specific scenarios.
To solve this problem, a branch of knowledge-enhanced PLMs~\cite{Zhang2019ERNIE, Wang2021KAdapter, Pan2022Knowledge} have been proposed for capturing rich factual knowledge from external knowledge bases. 
In addition, some recent large-scale PLMs (e.g., GPT-3~\cite{Brown2020Language}) can enable few/zero-shot in-context learning with language prompts or instructions. Thus, we can leverage cross-task learning to unify semantics knowledge from different NLP tasks.

\subsection{Fine-tuning for PLMs}
A large number of applications in real scenarios focus on how to fine-tune the PLM to transfer the prior knowledge derived from the general domain to downstream task-specific domains~\cite{Xu2020CLUE, Wang2019GLUE}.
We integrate some task-orient fine-tuning methods to allow users to develop and evaluate PLM on different NLP tasks.
We also implement some popular tuning algorithms to enable tuning on low-resource scenarios, such as prompt-tuning~\cite{Liu2021GPT}, in-context learning~\cite{Brown2020Language}, etc.

\begin{figure*}[th!]
    \centering
	\includegraphics[width=\linewidth]{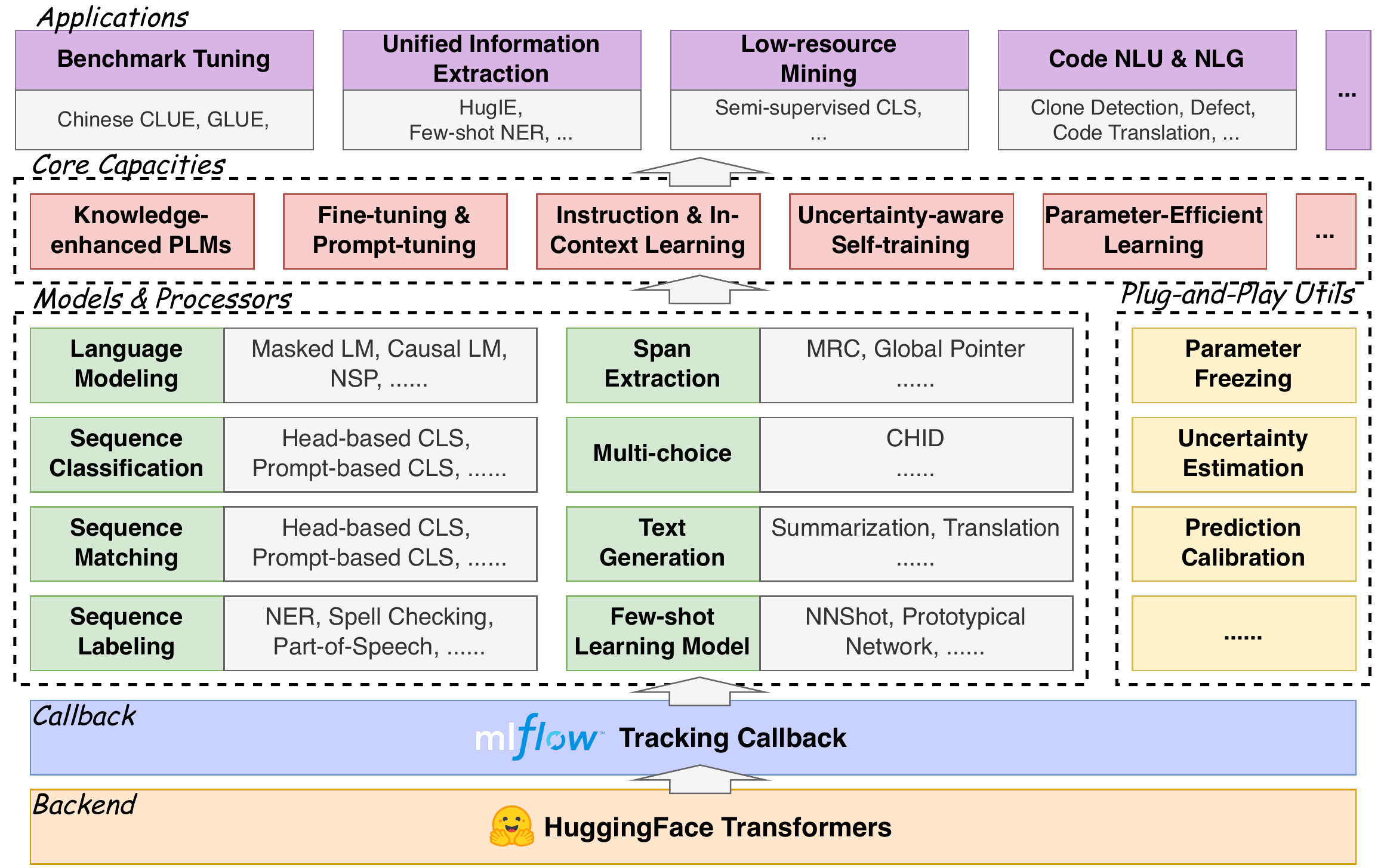}
	\caption{An overview of the {\model} library.}\label{fig:overview}
\end{figure*}

\section{{\model}}


\subsection{Overview}

{\model} is an open-sourced library with a hierarchical structure. As shown in Figure~\ref{fig:overview}. The backend is the prevalent HuggingFace Transformers platform that provides multiple transformer-based models and task trainers. In other words, {\model} can be seen as a customized NLP platform for efficiently training and evaluating.
In addition, {\model} integrates \emph{MLFlow}, which is a novel tracking callback toolkit for model training and experiment result analysis. Users can simply add one configure parameter \texttt{tracking\_uri} in the training script, and observe the tracking records after running \emph{MLFlow} server.

{\model} consists of three key components, including \emph{Models}, \emph{Processors}, and \emph{Applications}. Users can directly select the pre-built settings for some common tasks, or develop special user-defined training solutions in real-world application scenarios. We will provide a detailed description in the following sections.

\subsection{Library Architecture}

\paragraph{Models.} 

In \emph{Models}, we provide some popular transformer-based models as backbones, such as BERT, RoBERTa, GPT-2, etc. We also release our pre-built KP-PLM, a novel knowledge-enhanced pre-training model which leverages \emph{knowledge prompting}~\cite{Wang2022Knowledge} paradigm to inject factual knowledge and can be easily used for arbitrary PLMs.
Apart from basic PLMs, we also implement some task-specific models, involving sequence classification, matching, labeling, span extraction, multi-choice, and text generation.
Particularly, we develop standard fine-tuning (based on CLS Head~\footnote{For standard fine-tuning, we need to add a classification head (CLS head) on the PLM and obtain the probability distribution of each class. The parameters of the CLS head are randomly initialized.}) and prompt-tuning models~\footnote{Different from fine-tuning, prompt-tuning can reuse the pre-training objective (e.g., MLM, CLM) to perform classifying on the masked token. It requires a task-orient template (e.g., ``It was [MASK].'') and the label word mapping (e.g., ``great'' maps to ``positive'' class in sentiment analysis task.)} that enable PLM tuning on classification tasks.
For few-shot learning settings, {\model} provides a prototypical network~\cite{Snell2017Prototypical} in both few-shot text classification and named entity recognition (NER).

\begin{figure}
\begin{minipage}{0.48\textwidth}
\begin{lstlisting}[language=python, caption=A model case of parameter freezing., label=case1, frame=shadowbox]
from tools.model_utils.parameter_freeze import      
    ParameterFreeze
freezer = ParameterFreeze()
class BertForSequenceClassification(BertPreTrained
    Model):
    def __init__(self, config):
        super().__init__(config)
        self.num_labels = config.num_labels
        self.config = config
        self.bert = BertModel(config)
        # freeze the backbone
        if self.config.use_freezing:
            self.bert = freezer.freeze_lm(self.bert)
        self.classifier = torch.nn.Linear(
            config.hidden_size, config.num_labels)
        self.init_weights()
\end{lstlisting}
\end{minipage}
\end{figure}

In addition, we also incorporate some \textit{plug-and-play utils} in {\model}. 
1) \emph{Parameter Freezing}. If we want to perform parameter-efficient learning~\cite{Mao2022UniPELT}, which aims to freeze some parameters in PLMs to improve the training efficiency, we can set the configure \texttt{use\_freezing} and freeze the backbone. A use case is shown in Code~\ref{case1}.
2) \emph{Uncertainty Estimation} aims to calculate the model certainty when in semi-supervised learning~\cite{Mukherjee2020Uncertainty}.
3) We also design \emph{Prediction Calibration}, which can be used to further improve the accuracy by calibrating the distribution and alleviating the semantics bias problem~\cite{Zhao2021Calibrate}.

\paragraph{Processors.} 

{\model} aims to load the dataset and process the task examples in a pipeline, containing sentence tokenization, sampling, and tensor generation.
Specifically, users can directly obtain the data through \texttt{load\_dataset}, which can directly download it from the Internet or load it from the local disk.
For different tasks, users should define a task-specific data collator, which aims to transform the original examples into model input tensor features.

\paragraph{Applications.}
It provides rich modules for users to build real-world applications and products by selecting among an array of settings from \emph{Models} and \emph{Processors}. 
More details are shown in Section~\ref{sec:application}.

\subsection{Core Capacities}

To further improve the effectiveness of {\model}, we design multiple core capacities in the following.

\paragraph{Knowledge-enhanced Pre-training.}
Conventional pre-training methods lack factual knowledge ~\cite{Zhang2022DKPLM, Pan2022Knowledge}. 
To deal with this issue, 
we present KP-PLM~\cite{Wang2022Knowledge} with a novel knowledge prompting paradigm for knowledge-enhanced pre-training.
Specifically, we construct a knowledge sub-graph for each input text by recognizing entities and aligning with the knowledge base (e.g., Wikidata5M~\footnote{\url{https://deepgraphlearning.github.io/project/wikidata5m}.}) and decompose this sub-graph into multiple relation paths, which can be directly transformed into language prompts.
KP-PLM can be easily applied to other PLMs without introducing extra parameters as knowledge encoders.

\paragraph{Prompt-based Fine-tuning.}
Prompt-based fine-tuning aims to reuse the pre-training objective (e.g., MLM) and utilizes a well-designed template and verbalizer to make predictions, which has achieved great success in low-resource settings.
We integrate some novel approaches into {\model}, such as PET~\cite{Schick2021Exploiting}, P-tuning~\cite{Liu2021GPT}, etc.

\paragraph{Instruction-tuning and In-Context Learning.}
Instruction-tuning \cite{Wei2022Finetuned} and in-context learning~\cite{Brown2020Language} enable few/zero-shot learning without parameter update, which aims to concatenate the task-aware instructions or example-based demonstrations to prompt GPT-style PLMs to generate reliable responses.
So, all the NLP tasks can be unified into the same format and can substantially improve the models' generalization.
Inspired by this idea, we extend it into other two paradigms:
1) extractive-style paradigm: we unify various NLP tasks into span extraction, which is the same as extractive question answering~\cite{Keskar2019Unifying}, and 2) inference-style paradigm: all the tasks can be viewed as natural language inference to match the relations between inputs and outputs~\cite{Wang2021Entailment}.

\begin{figure}
\begin{minipage}{0.48\textwidth}
\begin{lstlisting}[language=python, caption=An application case of sequence classification for GLUE benchmark., label=case2, frame=shadowbox]
python3 hugnlp_runner.py \
  --model_name_or_path=$path \
  --data_dir=$data_path \
  --output_dir=./outputs/glue/$glue_task \
  --seed=42 \
  --max_seq_length=$len \
  --max_eval_seq_length=$len \
  --do_train \
  --do_eval \
  --per_device_train_batch_size=8 \
  --per_device_eval_batch_size=4 \
  --gradient_accumulation_steps=1 \
  --evaluation_strategy=steps \
  --learning_rate=1e-5 \
  --num_train_epochs=10 \
  --task_name=clue \
  --task_type=head_cls \
  --model_type=bert \
  --user_defined="data_name=rte" \
\end{lstlisting}
\end{minipage}
\end{figure}

\begin{figure}
    \centering	\includegraphics[width=\linewidth, frame]{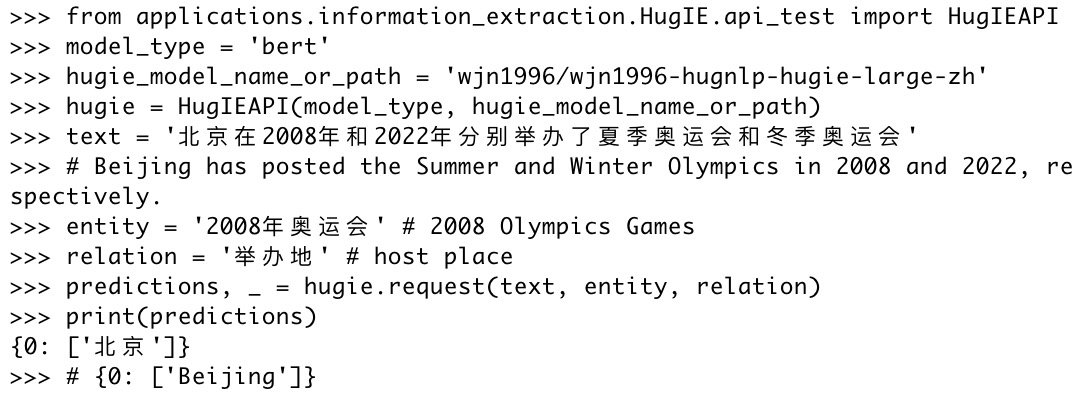}
	\caption{An application case of HugIE.}\label{fig:case3}
\end{figure}

\paragraph{Uncertainty-aware Self-training.}
Self-training can address the labeled data scarcity issue by leveraging the large-scale unlabeled data in addition to labeled data, which is one of the mature paradigms in semi-supervised learning~\cite{Qi2022Small, Nitesh2005Learning, Amini2022Self}.
However, the standard self-training may generate too many noises, inevitably degrading the model performance due to the confirmation bias.
Thus, we present uncertainty-aware self-training. Specifically, we train a teacher model on few-shot labeled data, and then use Monte Carlo (MC) dropout technique in Bayesian neural network (BNN)~\cite{Gal2016Dropout} to approximate the model certainty, and judiciously select the examples that have a higher model certainty of the teacher.

\paragraph{Parameter-efficient Learning.}
To improve the training efficiency of {\model}, we also implement parameter-efficient learning, which aims to freeze some parameters in the backbone so that we only tune a few parameters during model training.
We develop some novel parameter-efficient learning approaches, such as Prefix-tuning~\cite{Li2021Prefix}, Adapter-tuning~\cite{Houlsby2019Parameter}, BitFit~\cite{Zaken2022BitFit} and LoRA~\cite{Hu2022LoRA}, etc.

\begin{figure*}
    \centering
	\includegraphics[width=\linewidth]{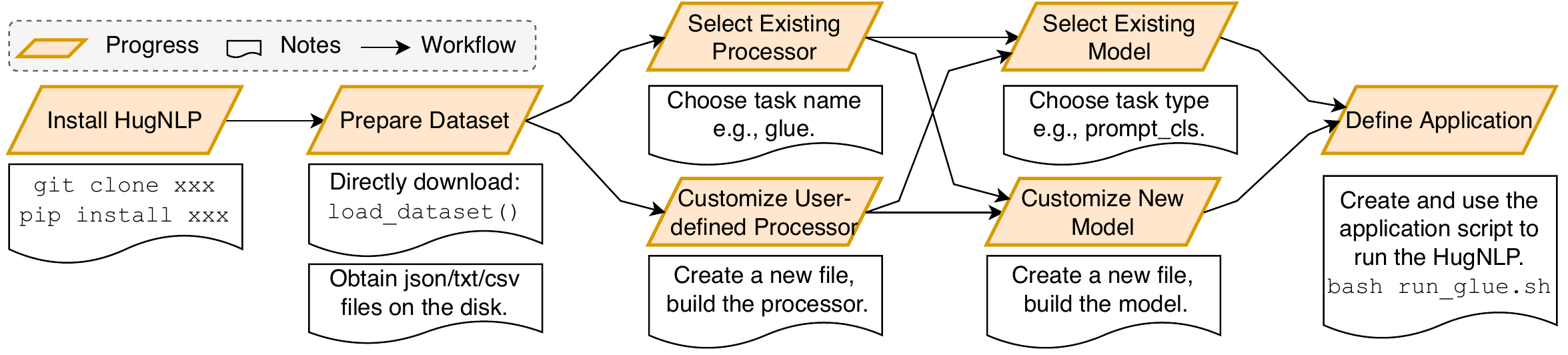}
	\caption{The development workflow of {\model}.}\label{fig:workflow}
\end{figure*}

\subsection{Featured Applications}
\label{sec:application}

\begin{table*}
 \centering
 \begin{small}
 \begin{tabular}{l cccccccc} 
  \toprule 
 \bf PLMs   &  \bf AFQMC & \bf CMNLI &  \bf CSL &  \bf IFLYTEK &  \bf OCNLI &  \bf TNEWS &  \bf WSC & \bf Avg. \\ 
  \midrule 
BERT-base & 72.30 & 75.91 & 80.83 & 60.11 & 78.52 & 57.18 & 75.89 & 72.04 \\
BERT-large & 72.91 & 77.62 & 81.30 & 60.77 & 78.71 & 57.77 & 78.28 & 72.60 \\
RoBERTa-base & 73.33 & 81.05 & 80.17 & 60.81 & 80.88 & 57.69 & 86.74 & 74.10 \\
RoBERTa-large & 74.66 & 80.50 & 82.60 & 61.37 & 82.19 & 58.54 & 87.53 & 75.33 \\
MacBERT-base & 74.23 & 80.65 & 81.63 & 61.14 & 80.65 & 57.65 & 80.26 & 73.80 \\
MacBERT-large & 74.66 & 81.19 & 83.70 & 62.05 & 81.92 & 59.03 & 86.74 & 75.46 \\
 \bottomrule 
 \end{tabular}
 \end{small}
\caption{Accuracy (\%) of different tasks in the CLUE benchmark.}
\label{table:benchmark-clue}
\end{table*}

\paragraph{Benchmark Tuning.}
We develop the training application for some popular benchmarks, such as Chinese CLUE and GLUE.
We use both standard fine-tuning and prompt-based fine-tuning paradigms to tune PLMs over these benchmarks. 
The case of this application is shown in Code~\ref{case2}.

\paragraph{Universal Information Extraction based on Extractive Instruction.}
We develop HugIE, a novel universal information extraction toolkit based on {\model}. Specifically, we collect multiple Chinese NER and event extraction datasets from ModelScope~\footnote{\url{https://modelscope.cn/datasets}} and QianYan~\footnote{\url{https://www.luge.ai}}. Then, we use the core capacity of extractive-style instruction with a global pointer~\cite{Su2022Su} to pre-train a universal information extraction model.
We also upload the trained model to HuggingFace~\footnote{\url{https://huggingface.co/wjn1996/wjn1996-hugnlp-hugie-large-zh}.}.
An example of using HugIE is shown in Figure~\ref{fig:case3}.

\paragraph{Low-resource Tuning for PLMs.}
For low-resource settings,
we have integrated two core capacities of prompt-tuning and uncertainty-aware self-training to further improve the performance with limited labeled data.
In other words, prompt-tuning can fully reuse the prior knowledge derived from PLMs to achieve high grades with few examples, while self-training can augment unlabeled data to enhance effectiveness.

\paragraph{Code Understanding and Generation.}
In addition to traditional NLP tasks, we also consider the scenario of code understanding and generation, 
such as clone detection, defect detection, and code summarization~\cite{Lu2021codexglue}.


\subsection{Development Workflow}

{\model} is easy to use and develop. We draw a workflow in Figure~\ref{fig:workflow} to show how to develop a new running task.
It consists of five main steps, including library installation, data preparation, processor selection or design, model selection or design, and application design.
This illustrates that {\model} can simplify the implementation of complex NLP models and tasks.

\begin{table*}[ht]
\centering
\resizebox{\linewidth}{!}{
\begin{tabular}{c  c  llllllll  c}
\toprule
\multirow{2}*{\bf Paradigms} & \multirow{2}*{\bf Methods} & \bf SST-2 & \bf SST-5 & \bf MR & \bf CR & \bf MPQA & \bf Subj & \bf TREC & \bf CoLA & \multirow{2}*{\bf Avg.} \\
&  & (acc) & (acc) & (acc) & (acc) & (acc) & (acc) & (acc) & (matt.) & \\
\midrule
\multirow{2}*{PT-Zero} & RoBERTa & 82.57	& 29.46 & \textbf{65.10} & \textbf{82.15} & 49.90 & \textbf{69.20} & 20.80 & -4.89 & 49.29 \\
& KP-PLM & \textbf{84.15} &	\textbf{30.67} &	64.15 &	81.60 &	\textbf{53.80} &	68.70 &	\textbf{24.80} &	\textbf{-2.99} &	\textbf{50.61} \\
\hdashline
\multirow{2}*{PT-Few} & RoBERTa & 86.35\small{\textpm1.3} &	36.79\small{\textpm2.0} &	\textbf{83.35}\small{\textpm0.9} &	\textbf{88.85}\small{\textpm1.4} &	66.40\small{\textpm1.9} &	89.25\small{\textpm2.6} &	76.80\small{\textpm5.0} &	6.61\small{\textpm6.9} & 66.80 \\
& KP-PLM &	\textbf{90.71}\small{\textpm1.0} &	\textbf{44.21}\small{\textpm2.9} &	82.00\small{\textpm1.5} &	85.35\small{\textpm0.4} &	\textbf{67.30}\small{\textpm1.2} &	\textbf{91.45}\small{\textpm0.4} &	\textbf{81.00}\small{\textpm3.3} &	\textbf{24.28}\small{\textpm11.3} &	\textbf{70.79} \\
\hdashline
\multirow{2}*{FT-Full} & RoBERTa &	94.90 &	56.90 &	\textbf{89.60} &	88.80 &	86.30 &	\textbf{96.50} &	\textbf{97.10} &	63.90 &	84.25 \\
& KP-PLM &	\textbf{95.30} &	\textbf{57.63} &	89.20 &	\textbf{89.10} &	\textbf{87.40} &	96.20 &	\textbf{97.10} &	\textbf{64.87} &	\textbf{84.60} \\

\bottomrule
\end{tabular}
}
\caption{The comparison between KP-PLM and RoBERTa-base over multiple natural language understanding (NLU) tasks in terms of acc/f1/matt. (\%) and standard deviation with three paradigms, such as zero-shot prompt-tuning (PT-Zero), few-shot prompt-tuning (PT-Few), and full-data fine-tuning (FT-Full).}
\label{tab:nlu}
\end{table*}

\begin{table*}[t]
\begin{center}
\resizebox{\textwidth}{!}{
\begin{tabular}{p{12cm}p{6cm}}
\begin{minipage}[t]{0.775\textwidth}
\resizebox{\textwidth}{!}{
\begin{tabular}{lc c  c c  c c  c c c}
\toprule
\multirow{2}{*}{\bf Methods} & \multirow{2}{*}{\bf Params.} & \multicolumn{2}{c}{\bf Java to C\#} & \multicolumn{2}{c}{\bf C\# to Java}  & \multicolumn{2}{c}{\bf Refine Small}  & \multicolumn{2}{c}{\bf Refine Medium}\\ 
\cmidrule(lr){3-4}\cmidrule(lr){5-6}\cmidrule(lr){7-8}\cmidrule(lr){9-10}
&  & (bleu) & (em)  & (bleu) & (em) & (bleu) & (em) & (bleu) & (em)  \\ 
\midrule
\textbf{CodeT5} &  &  &  &  &  &  &  &  &  \\
Fine-Tuning & 224M \cellcolor[gray]{.5}& \textbf{84.15} & \textbf{65.30} & \textbf{79.12} & \textbf{66.40} & 77.39 & \textbf{21.35} & \textbf{91.04} & \textbf{7.82} \\
BitFit & 0.001M \cellcolor[gray]{.9}& 0.25 & 0.00 & 0.24 & 0.00 & 1.28 & 0.00 & 5.14 & 0.00 \\
Adapter & 14.22M \cellcolor[gray]{.7}& 75.43 & 52.40 & 73.10  & 57.70 & 77.41 & 18.58 & 91.01 & 3.61 \\
P-Tuning V2 & 0.633M \cellcolor[gray]{.8}& 59.86 & 33.70 & 57.10 & 41.00 & \textbf{78.99} & 4.56 & 91.02 & 0.79 \\
\cdashline{1-10}
\\[-1em]
\textbf{PLBART} &  &  &  &  &  &  &  &  \\
Fine-Tuning & 139M \cellcolor[gray]{.55}& \textbf{77.05} & \textbf{62.60} & \textbf{79.29} & \textbf{62.80} & 73.32 & \textbf{12.71} & 83.88 & \textbf{4.24} \\
BitFit & 0.126M \cellcolor[gray]{.89}& 16.48 & 0.10 & 17.43 & 0.90 & \textbf{74.08} & 1.45 & \textbf{85.41} & 0.42 \\
Adapter & 7.11M \cellcolor[gray]{.76}& 66.72 & 42.10 & 68.70 & 51.00 & 73.58 & 10.90 & 84.72 & 3.12 \\
P-Tuning V2 & 0.329M \cellcolor[gray]{.87}& 22.87 & 1.00 & 48.08 & 33.80 & 73.87 & 2.07 & 73.58 & 0.03 \\
\bottomrule
\end{tabular}
}
\captionsetup{type=table}
\vspace{-0.5em}  
\caption{Performance (\%) on Code Translation \& Code Refinement Tasks.}
\label{table:code-trans}

\end{minipage}

&

\begin{minipage}[t]{0.37\textwidth}
\resizebox{\textwidth}{!}{
\begin{tabular}{lccc}
\toprule
\multirow{2}{*}{\bf Methods} & \multirow{2}{*}{\bf Params.} & \bf Defect & \bf Clone  \\ 
\cmidrule(lr){3-3}\cmidrule(lr){4-4} 
&  & (acc) & (f1) \\
\midrule
\textbf{CodeT5} & &  &  \\
Fine-Tuning & 224M \cellcolor[gray]{.5}& \textbf{64.35} & \textbf{94.97} \\
BitFit &1.183M \cellcolor[gray]{.84}& 55.05 & 69.52 \\
Adapter & 15.40M \cellcolor[gray]{.68}& 59.74 & 94.47 \\
P-Tuning V2 & 1.182M \cellcolor[gray]{.84}& 54.61 & 79.83 \\
\cdashline{1-4}
\\[-1em]
\textbf{PLBART} &  &  &  \\
Fine-Tuning & 139M \cellcolor[gray]{.55}& \textbf{62.27} & \textbf{92.85} \\
BitFit & 1.308M \cellcolor[gray]{.83}& 56.30 & 92.42 \\
Adapter & 8.29M \cellcolor[gray]{.74}& 61.60 & 92.74 \\
P-Tuning V2 & 1.182M \cellcolor[gray]{.84}& 53.81 & 75.88 \\

\bottomrule
\end{tabular}
}
\captionsetup{type=table}
\vspace{-0.5em}
\caption{Performance (\%) on Code Clone Detection \& Code Defect Detection Tasks.}
\label{table:code-understanding}
\end{minipage}
\end{tabular}
}
\end{center}
\end{table*}

\section{Experimental Performances}
In this section, we empirically examine the effectiveness and efficiency of the {\model} toolkit on some public datasets.

\subsection{Performance of Benchmarks}

To validate the effectiveness of {\model} on both fine-tuning and prompt-tuning, we choose Chinese CLUE~\cite{Xu2020CLUE} and GLUE benchmarks~\cite{Wang2019GLUE}. 
For Chinese CLUE, we choose different sizes of BERT, RoBERTa and MacBERT~\cite{Cui2020Revisiting} and report the accuracy over the development sets of each task in Tables~\ref{table:benchmark-clue}.
For GLUE, we perform full-resource fine-tuning (FT-full), few-shot prompt-tuning (PT-few), and zero-shot prompt-tuning (PT-zero) based on our proposed KP-PLM. We select RoBERTa as the strong baseline and report the accuracy results with standard deviation in Table~\ref{tab:nlu}.
The obtained comparable performance has shown the reliability of {\model} in both full and low-resource scenarios, which achieves similar performance compared to other open-source frameworks and their original implementations~\cite{Wang2022EasyNLP}.

\subsection{Evaluation of Code-related Tasks}
We use {\model} to evaluate the performance on multiple code-related tasks, such as code clone detection, defection, translation, and refinement. We fine-tune two widely used models: CodeT5~\cite{Wang2021CodeT5} and PLBART~\cite{ahmad2021unified}, and then compare them with competitive  parameter-efficient learning methods, including BitFit, Adapter, and P-tuning V2~\cite{Liu2021Ptuningv2}. 
Results in Table~\ref{table:code-trans} and Table~\ref{table:code-understanding} demonstrate the effectiveness and efficiency of {\model}.

\begin{table}
 \centering
 \resizebox{\linewidth}{!}{
 \begin{tabular}{l cccc} 
  \toprule 
 \bf Methods   &  \bf RTE & \bf CB &  \bf AGNews & \bf Avg. \\ 
  \midrule 
\multicolumn{5}{l}{\textit{\textbf{Few Labeled Data (16-shot)}}}\\
Fine-Tuning & 54.4\small{\textpm3.9} & 74.5\small{\textpm2.6} & 88.9\small{\textpm2.7} & 72.60 \\
\midrule
\multicolumn{5}{l}{\textit{\textbf{Few Labeled Data (16-shot) + Unlabeled Data}}}\\
UST & 55.6\small{\textpm2.6} & 76.0\small{\textpm3.1} & 89.3\small{\textpm3.5} & 73.63 \\
CEST & 57.0\small{\textpm1.9} & 78.1\small{\textpm2.7} & 88.5\small{\textpm2.2} & 74.53 \\
LiST & \bf 60.8\small{\textpm2.5} & \bf 79.7\small{\textpm2.9} & \bf 90.3\small{\textpm2.5} & \bf 76.93 \\
 \bottomrule 
 \end{tabular}
 }
\caption{Accuracy (\%) of uncertain-aware self-training with only 16 labeled examples per class.}
\label{table:self-training}
\end{table}

\subsection{Effectiveness of Self-training}
We end this section with an additional validation on the self-training. We choose some recent methods (using uncertainty estimation) to evaluate the implementations of {\model}, including UST~\cite{Mukherjee2020Uncertainty}, CEST~\cite{Tsai2022Contrast}, and LiST~\cite{Wang2022LiST}.
Results in Table~\ref{table:self-training} show that self-training can make substantial improvements in low-resource scenarios.

\section{Conclusion}
In this paper, we introduce {\model}, a unified and comprehensive library based on PyTorch and HuggingFace, allowing researchers to apply it to different academics and industry scenarios.
{\model} consists of three key components (i.e., \emph{Processors}, \emph{Models} and \emph{Applications}) and multiple pre-built core capacities and plug-and-play utils.
Finally, we perform some evaluation of different aspects of applications, and the results demonstrate its efficiency and effectiveness. We think {\model} can promote research and development for NLP applications.

\section*{Ethics Statement}
Our contribution in this work is to construct a unified and comprehensive library for NLP research and application.
However, transformer-based models may have some negative impacts, such as gender and social bias. 
Our work would unavoidably suffer from these issues.
We suggest that users should carefully address potential risks 
when models trained using the {\model} library are deployed online.

\section*{Acknowledgements}
This work has also been supported by the National Natural Science Foundation of China under Grant No. U1911203, 
Alibaba Group through the Alibaba Innovation Research Program, 
and the National Natural Science Foundation of China under Grant No. 61877018,
the Research Project of Shanghai Science and Technology Commission (20dz2260300) and the Fundamental Research Funds for the Central Universities.

\bibliography{custom}
\bibliographystyle{acl_natbib}




\end{document}